\documentclass[10pt,twocolumn]{article}
\usepackage{graphicx}
\usepackage{amsmath,amssymb}

\newcommand{\tmem}[1]{{\em #1}}
\newtheorem{definition}{Definition}
\newcommand{\tmstrong}[1]{\textbf{#1}}
\newcommand{\tmop}[1]{\operatorname{#1}}

\newtheorem{theorem}{Theorem}
\newtheorem{lemma}{Lemma}

\newcommand{\Section}[1]{\vspace{-8pt}\section{\hskip -1em.~~#1}\vspace{-3pt}}

\begin{document}

\date{November 2003\footnote{This paper was originally written in November 2003,
but has been submitted to Arxiv in 2007.  References have not been updated to
include more recent work.}
}

\title{\Large\bf View Based Methods can achieve Bayes-Optimal 3D Recognition}
\author{Thomas M. Breuel\\PARC, Palo Alto, USA\\{\tt tmb@parc.com}}

\maketitle

\section*{\centering Abstract}
\textit{This paper proves that visual object recognition systems using only 2D
Euclidean similarity measurements to compare object views against previously
seen views can achieve the same recognition performance as 
observers having access to all coordinate information and able of using
arbitrary 3D models internally. 
Furthermore, it demonstrates that such systems do not require more training views
than Bayes-optimal 3D model-based systems.  For building computer vision
systems, these results imply that using view-based or appearance-based
techniques with carefully constructed combination of evidence mechanisms may
not be at a disadvantage relative to 3D model-based systems.  For
computational approaches to human vision, they show that it is impossible to
distinguish view-based and 3D model-based techniques for 3D object
recognition solely by comparing the performance achievable by human and 3D
model-based systems.}

\Section{Introduction}

View-based or appearance based methods in visual object recognition represent
3D objects as a collection of views for the purposes of recognition.  Many
different ways in which these views can be used for recognition have been
proposed: some compare a target view against stored views individually, while
others allow interpolation or combination among multiple views.  Some
approaches use fixed similarity functions and evidence combination schemes,
while others allow for the learning or adaptation of either or both.

One of the most restrictive forms of view-based 3D object recognition requires
that, in order to perform recognition, each stored view is compared with a
target view using only a fixed, non-invariant similarity measure.  After
performing those similarity measurements, the observer is then permitted to
perform some kind of ``combination of evidence'' on them.  In their papers on
human 3D generalization {\cite{LiuKniKer95}}{\cite{LiuKer98}} refer to such an observer
as an observer using a {\tmem{strong view-approximation method}}: 

\begin{quotation}
{\it
  \noindent ``For example, assume that an object is represented by two independent views.
  The task is to decide whether a novel view belongs to the object. The strong
  version of view-approximation maintains that in order to recognize a novel
  view, a similarity measure is calculated independently between this view and
  each of the two stored views [...]. Recognition is a function of these
  measurements. The simplest function is the nearest neighbor scheme, where a
  match is based on the closest view in memory. A more sophisticated scheme is
  the Bayes classifier that combines the evidence over the collection of views
  optimally.'' \cite{LiuKer98}
}
\end{quotation}

Let us express this notion of ``strong view-approximation'' formally.  We will
call an observer using a strong version of the view-approximation
method{\footnote{The same paper defines a supposedly ``more flexible'' version
of view approximation, in which the observer is permitted to perform geometric
transformations on the target or training view.  Since we demonstrate in this
paper that the strong view-approximation method is already sufficient for
achieving Bayes-optimal 3D performance, we need not consider ``more flexible''
models.}} a ``strongly two-dimensional observer'':

\begin{definition}\label{strict}
  Let $\mathfrak{T} = \{ T_{\omega, i} : \omega \in \Omega, i = 1, \ldots,
  r_{\omega} \}$ be a collection of $N$ 2D training views $T_{\omega, i}$ for
  objects $\omega \in \Omega = \{ \omega_1, \ldots, \omega_N \}$.  Let $S ( U,
  V )$ be a real-valued function of 2D views, the {\em view similarity measure}.
  Then, a {\tmstrong{strongly
  two-dimensional observer}} is an observer that classifies an unknown target
  view $V$ using a decision procedure $D ( V )$ of the form
  \[ D_{} ( V ) = f ( S ( V, T_{\omega_1, 1} ), S ( V, T_{\omega_1, 2} ),
     \ldots, S ( V, T_{\omega_N, r_N} )) \]
  That is, a strongly two-dimensional observer classifies objects only based
  on some functional combination $f$ of the individual 2D similarities of the
  target view to each of the training views.
\end{definition}

Note that the observer is permitted to take into account in his
decision similarities to both matching and non-matching
objects{\footnote{However, while this perhaps the most plausible
definition, the results of this paper do not depend on it; see
Appendix B.}}.  For example, in nearest neighbor methods, we compare
similarities from both matching and non-matching objects in order to
find the view having the highest similarity value (i.e., smallest
Bayes-optimal distance).

Intuitively, it would seem that a strongly two-dimensional observer
should be limited in his ability to perform recognition and should
therefore make more recognition errors than an observer capable of
performing full, 3D modeling and recognition.  In this paper, I
demonstrate that that is not the case: given the correct Bayesian
combination of the individual view similarity values, a strongly
two-dimensional observer can achieve the same Bayes-optimal error rate
as an observer that can access all the coordinate measurements of the
target and training views and uses explicit 3D models internally.
This is demonstrated by showing that an observer can reconstruct the
original training and target views well enough from the similarity
values to be able to perform Bayes-optimal 3D recognition.
Furthermore, I show that the same result holds true for model
acquisition: a strongly two-dimensional observer can acquire object
models just as quickly and reliably from view similarity values as an
observer having full access to views.

\Section{Bayes-Optimal 3D Recognition}

Assume that we are trying to identify which of a number of possible objects
$\omega$ is represented by some view $V$ of the object.  The Bayes-optimal
minimum error decision procedure $D ( V )$ for this problem is to determine the
object with the largest posterior probability given the image:
\[ D ( V ) = \tmop{arg} \max_{\omega} P ( \omega | V ) \]
Via Bayes rule, we can compute $P ( \omega | V )$ in terms of the likelihood
$P ( V | \omega )$:
\[ P ( \omega | V ) = \frac{P ( V | \omega ) P ( \omega )}{P ( V )} \]
Since $P ( V )$ is independent of the object, our decision procedure then simply
becomes
\[ D ( V ) = \tmop{arg} \max_{\omega} P ( V | \omega ) P ( \omega ) \]

Now, let $M_{\omega}$ be the true 3D model corresponding to object $\omega$,
let $R$ be the 3D object transformation and imaging transformation, and let
$N$ be the noise and uncertainty introduced by the imaging process.  Then, the
target view $V$ is distributed as
\[ V \sim R ( M_{\omega} ) + N \text{} \]
Here, $M_{\omega},$ $R$, and $N$ are all random variables.  In different
words, we can write down a conditional distribution of $V$ given $R$,
$M_{\omega}$, and $N$.  However, $R$ and $N$ are unobservable.  Hence, a
Bayes-optimal 3D observer needs to take into account his prior knowledge about
the distribution of those variables to arrive at an expression for $P ( V |
\omega$):
\[ P_M ( V | \omega ) = P ( V | M_{\omega} ) = \int P ( V | M_{\omega}, R, N )
   P ( R, N | M_{\omega} ) d N d R \]
Note that we allow both the distribution of noise $N$ and the distribution of
views $R$ to depend on the model; commonly (though not necessarily correctly),
it is assumed that these are independent, so that $P ( R, N | M_{\omega} ) = P
( R ) P ( N )$.

By construction, an observer using $P_M ( V | \omega$) is using the
{\em Bayes-optimal object recognition procedure for 3D objects from 2D
views} and achieves the Bayes-optimal error rate on the recognition
problem given $M_{\omega}$.

In actual practice, an observer almost never knows the true 3D object model
$M_{\omega}$, but needs to reconstruct it from a given set of training views
$\mathcal{T}_{\omega} = \{ T_{\omega, 1}, \ldots, T_{\omega, r} \}$.  In
general, the 3D model cannot be reconstructed unambiguously from the training
views, due to noise, uncertainty, ambiguity, and/or occlusions.  Therefore,
the observer really can only estimate a distribution $P ( M_{\omega} |
\mathcal{T}_{\omega} )$ and the actual model also becomes a latent variable:
\begin{eqnarray}\label{opt3d}
  P_T ( V | \omega ) &=& P ( V | \mathcal{T}_{\omega} ) \\
     &=& \int P ( V |
  M_{\omega}, R, N ) P ( R, N | M_{\omega} ) \nonumber \\
	&\,&\;\;\;\;\cdot P ( M_{\omega} |
	  \mathcal{T}_{\omega} )\,dN\,dR\,dM_{\omega} \nonumber
\end{eqnarray}
An observer using $P_T ( V | \omega )$ is the {\em Bayes-optimal 3D
observer based on a set of 2D training views} and achieves minimum
recognition error for the given prior distributions.

The difference between $P_M ( V | \omega )$ and $P_T ( V | \omega )$ is
crucial: an observer having {\tmem{a priori}} knowledge of the correct 3D
structure $M_{\omega}$ of object $\omega$ can easily outperform an observer
who has to estimate such a model from training views $\mathcal{T}_{\omega}$. 
However, where would an observer obtain exact knowledge of $M_{\omega}$?  The
observer might have access to information beyond a set $\mathcal{T}_{\omega}$
of given training views, such as information derived from touch or a given
CAD (computer-aided design) blueprint; but then we are comparing the
performance of view-based recognition against the performance of an observer
that has additional information.

The observer might also try to perform an ``optimal reconstruction''
$\hat{M}_{\omega}$ of $M_{\omega}$ based on $\mathcal{T}_{\omega}$
(e.g., using a maximum-likelihood procedure, maximum a posteriori--MAP, or
least-square reconstruction) and use that for matching; but that would
merely amount to picking $P ( M_{\omega} | \mathcal{T}_{\omega} ) =
\delta ( M_{\omega}, \hat{M}_{\omega} )$, which is almost certainly
not the correct distribution and would in general result in worse
performance than the Bayes optimal solution using the correct
distribution $P ( M_{\omega} | \mathcal{T}_{\omega} )$; we will return
to this issue below.

Therefore, the question of whether strongly view-based recognition
performs worse than a 3D recognition system only makes much sense if
we give both methods the same input data.  In the case of 3D
model-based recognition, we expect that the 3D model-based observer
should perform Bayes-optimal reconstruction of the 3D models
compatible with the training views, resulting in a distribution $P (
M_{\omega} | \mathcal{T_{\omega} )}$, and then would use that
distribution of models for recognition, as described by Equation
\ref{opt3d}

Note that we have, so far, not made any assumptions about the
representation of models or views; the above expressions are true for
collections of point features as much as they are true for grayscale
images.  However, it is common in the literature
{\cite{UllBas91}}{\cite{PogEde90}}{\cite{LiuKniKer95}}{\cite{LiuKer98}}
to examine the special case in which images are ordered collections of
$k$ points in $\mathbb{R}^2$, for some fixed $k$, models are
correspondingly ordered collections of $k$ points in $\mathbb{R}^3$,
noise $N$ has a Gaussian distribution around each image point, and
transformations consist of 3D rotations followed by orthographic
projection. For this formalization of the 3D object
recognition problem, views are vectors in $\mathbb{R}^{2 k}$ and
models are vectors in $\mathbb{R}^{3 k}$.  For concreteness and for a
connection with prior work, we use the same representation when
talking about a concrete instance of the recognition problem.
However, the derivations go through for other kinds of representations
and depend only on the use of Euclidean distances of views represented
as vectors\footnote{
Note particular that choosing to represent views as vectors in
$\mathbb{R}^n$ does not imply knowledge of feature correspondences;
for example, even if a view $V\in\mathbb{R}^{2k}$ represents the 2D
coordinates of feature points in that view, they might simply be
ordered lexicographically.  A representation of the input image
as a feature map or image also does not convey any feature correspondence
information.
}
in $\mathbb{R}^n$.

\Section{View-Based Recognition}

Let us now show that strong view-approximation methods can achieve
Bayes-optimal 3D recognition performance for feature-based object
recognition.  In fact, for our construction, we assume that the fixed
similarity measure used by the strong view-based approximation method
is simply the Euclidean distance.  That is, we define the similarity
function for a view $V$ and a view $T$ as $S ( V, T ) = \| V - T \|$.
Note that in the case where $V$ and $T$ are concatenations of the
locations of feature points in the image and the training view, this
is the same as a point wise squared error evaluation, $\sqrt{\sum_i
(v_i-t_i)^2}$, where the $v_i,t_i\in\mathbb{R}^2$ are corresponding
feature locations in the two vectors.

When attempting view-based recognition, we are comparing our unknown image $V$
against many previously stored training views $\mathfrak{T} = \bigcup
\mathcal{T}_{\omega} = \{ T_{\omega, i} : \omega \in \Omega, i = 1, \ldots,
r_{\omega} \} \subseteq \mathbb{R}^{2 k}$. We call this entire
collection $\mathfrak{T}$ of training views and their associated
object labels the {\tmem{model base}}.

When attempting to recognize an object from one of its views $V$, a
strongly two-dimensional view-based observer may take into account the
real-valued similarity of the view to each of the training views $S (
V, T_{\omega, i} )$ and combine them in some way.  The strongly
view-based observer is not permitted to evaluate $S$ for different
transformations of the views, or to perform calculations involving the
coordinates of the views, or perform any of the other operations that
model-based or view-based recognition systems commonly perform (e.g.,
\cite{Grimson90z}, \cite{belongie01shape}).

The definition of a strongly two-dimensional observer stated
informally by \cite{LiuKer98} and restated formally above does not
impose any restrictions on the kinds of knowledge an observer has
about the models in the model base, or the kinds of computations an
observer may perform on those models.  However, since we think of
visual systems as operating on-line and acquiring models
incrementally, we impose here the further restriction on the strongly
two-dimensional observer that his entire knowledge about the object in
the model base is limited to knowledge about their pairwise
similarities $S (T_{\omega, i}, T_{\omega', j} )$.  This strengthens
the result because it shows that an observer having even less
information than that required by the definition of the strongly
two-dimensional view-based observer can still perform Bayes-optimal 3D
recognition.

Let us call this entire collection of similarity measurements between
the training view and the views in the model base, together with the
pairwise similarities of views in the model base, $\mathfrak{S} ( V,
\mathfrak{T} )$.
A Bayes-optimal observer will combine them in a Bayes-optimal way.  We
will show the following theorem:

\begin{theorem}\label{main}
  Let $V$ and $T_{\omega, i}$ be object views represented as vectors
  in $\mathbb{R}^{2k}$. The collection of Euclidean similarity
  measurements $S ( V, T_{\omega, i} )$ against almost any model base
  of size $N \geq 2k$ is sufficient for performing Bayes-optimal
  3D recognition.
\end{theorem}

To show this, we will show that an observer can reconstruct the
$V$ and $T_{\omega,i}$ given $\mathfrak{S} ( V, \mathfrak{T} )$, up to
transformations that do not affect classification.
To establish this, we use the following Lemma:

\begin{lemma}
  For a collection of $N$ distinct vectors $p_1, \ldots, p_N$ that span
  $\mathbb{R}^n$, if $N \geq n$, we can reconstruct the coordinates of the
  vectors from the collection of Euclidean distances $d_{i j} = \| p_i - p_j
  \|$ up to a global translation, a global rotation, and mirror
  reversal.\label{recon}\label{reconstruction}
\end{lemma}

{\smallskip\noindent\it Proof.} See Appendix A.

{\smallskip\noindent\it Proof of Theorem~\ref{main}.} As defined
above, the target view
$V$ and each of the N training views $T_{\omega, i}$ is represented as
a point in $\mathbb{R}^{2 k}$.  We identify $n = 2 k$.  Furthermore,
we have the set of similarity measurements $\mathfrak{S} ( V,
\mathfrak{T} )$.  We identify the similarity measurements comparing
only the $N$ views in the model base with the $d_{i j}$ in the Lemma.
Lemma~\ref{recon} tells us that if the model base contains at least $2
k$ training views, then we can reconstruct the model base and the
target view from those similarity measurements, up to a single global
transformation $G$ (translation, global rotation, and mirror
reversal), provided that the set of training views spans
$\mathbb{R}^{2 k}$.

This will be true for {\tmem{almost all}} collections of $N$ training
views, for the following reason.  Consider the concatenation of the
$N$ training views into a vector ${\bf p}$ in the space of $N$
$n$-dimensional vectors, i.e., $\mathbb{R}^{N\cdot n}$.  This
collection of vectors can fail to satisfy the requirements of
Lemma~\ref{recon} either by not spanning $\mathbb{R}^n$ or by having
two vectors be identical.  Either of these is easily seen to constrain
${\bf p}$ to lie on a submanifold of $\mathbb{R}^{N\cdot n}$ of
measure zero.  Since there is only a finite number those constraints,
their union still has measure zero.

If we have some procedure for inferring $G$, then the proof is done at
this point: we can compute $G^{-1}$, reconstruct the target view $V$
and the set of training views $\mathfrak{T}$ exactly by first applying
Lemma~\ref{recon} and then transforming with $G^{-1}$, and finally
perform Bayes-optimal 3D recognition as defined by
Equation~\ref{opt3d}. This is a construction of a Bayes-optimal 3D
recognition procedure conforming to the requirements of
Definition~\ref{strict}.

For completeness, however, let us assume that $G$ cannot be determined
but that object identity is invariant under a global translation,
rotation, and mirror reversal transformation $G$ of both the target
views and all the training views.  This means that, for all target
views $V$ and sets of training views $\mathfrak{T}$, our decision
procedure $D$ is invariant under $G$:
\begin{equation}
D(V,\mathfrak{T}) = D(GV,G\mathfrak{T})
\end{equation}
Here $G\mathfrak{T} = \{GT | T\in\mathfrak{T}\}$.  If we apply
Lemma~\ref{recon}, it will reconstruct for us $GV$ and $G\mathfrak{T}$
for some such (unknown) transformation $G$.  But since, by assumption,
$D(V,\mathfrak{T}) = D(GV,G\mathfrak{T})$, if we apply our regular
decision procedure to the transformed training and target views, we
will be making the same decisions as if we had applied them to the
original training and target views.
Since Bayes-optimal 3D recognition, as expressed in
Equation~\ref{opt3d}, is a decision procedure of this form, it can be
evaluated in this way and will yield the same results on the target
and training views reconstructed from the similarity values as it does
on on the original target and training views.

Hence, by first reconstructing the target and training views using
Lemma~\ref{recon} and then applying Equation~\ref{opt3d}, we have
constructed a Bayes-optimal 3D recognition procedure using only 2D
similarity measurements between target and training views, as required
by Definition~\ref{strict}.  $\Box$

Before continuing, we should note that the appearance of the global
transformation $G$ is simply an artifact of the use of Euclidean distance
as our similarity measure, since Euclidean distances are invariant
under this set of transformations.  If we pick a similarity measure
that is not invariant, the uncertainty about $G$ disappear.
Appendix~B contains such a similarity measure.

The reason for using Euclidean distance in these derivations is that
it is, at the same time, an intuitive similarity measure for
similarity of 2D views and that the proof of
Lemma~\ref{reconstruction} is fairly easy.  The rotational invariance,
for example, can be eliminated by choosing a slightly more complicated
similarity function $S(V,T) =
\sqrt{\sum_i i\cdot(V_i-T_i)^2}$, but the analogous proof for
Lemma~\ref{reconstruction} becomes more complicated.

However, the appearance of $G$ is not a particularly serious issue.
If, in addition to the set of similarities, we know the actual 2D
coordinates of features in $2k+1$ training views (for example, from
tactile input), after applying Lemma~\ref{reconstruction} to obtain
$GV$ and $G\mathfrak{T}$, we can use those to determining $G^{-1}$ and
reconstruct the target view $V$ and training views exactly.  Note that
Definition~\ref{strict} permits such information to be available even
to a strictly two dimensional view based observer.

Another way of looking at this is that $G$ does not affect how we
measure translation and rotation of different views relative to each
other.  That is, informally stated, once we have decided that a
certain view represents, for example, ``vertical'', we can determine
the orientation of other views relative to that view even if we don't
know $G$.  That situation is somewhat analogous to phenomena observed
in human vision, which allow fairly rapid global reinterpretation of
globally transformed visual inputs \cite{Helmholtz25}; it is equivalent
to saying that $G$ remains unknown but that our decision procedure
is invariant under $G$, as in the second part of the proof above.

\paragraph{Note on Model Acquisition.}

The reader should recognize that the ``reconstruction'' of coordinates
from similarity measurements is a completely separate computation from
the acquisition of 3D models from 2D views (e.g.,
\cite{higgins81}).  The reconstruction above is concerned with the
recovery of $2k$-dimensional vectors from internally computed
similarity values among $2k$-dimensional vectors.  In 3D model
acquisition from 2D views, we attempt to combine views of an object,
possibly subject to sensor noise, into a consistent model.  3D model
acquisition could be carried out after the coordinates of the individual
views of an object have been reconstructed from similarity
measurements using the above procedure.

\paragraph{Other Feature Vectors.}

The same construction as described above applies to many other feature
types and situations, like grayscale or color images, feature locations
without correspondences, etc.

For example, if correspondences between feature locations image and
training views are not known, we can still concatenate the $k$
two-dimensional coordinates of those feature locations in each view
into a single vector in some arbitrary order and compute similarity,
as before, using Euclidean distances.  The resulting view similarity
measure would not be particularly nicely behaved, but it would still
satisfy the criteria of a strongly view based observer.  For
recognition using those similarity measures, the observer would
reconstruct the $2k$-dimensional vectors as before and then would have
to use some other method to find correspondences between different
views, just as if the observer had been given the original visual
input instead of similarities.

\paragraph{Actual Implementations.}

While the proof of the statistical sufficiency of $\mathfrak{S} ( V,
\mathfrak{T} )$ has involved the reconstruction of views from similarity
measurements, this is merely a mathematical device; it does not mean that
every Bayes-optimal view-based recognition system actually has to carry out
such a reconstruction.  Quite to the contrary, given a collection of millions
of stored training views $\mathfrak{T}$, it seems quite plausible that even
very simple decision functions, perhaps even something as simple as a linear
discriminant function on some fixed function $g$ of the similarity values,
$\Phi_{\omega} ( V ) = \sum_i \alpha_{\omega, i} g ( S ( V, T_{\omega, i} ))$,
may already represent a close approximation to the Bayes optimal error rate
and can be expected to converge to the Bayes-optimal 3D recognition error rate
for large enough sets of training views.  Note, in particular, that Radial
Basis Functions (RBFs) are of this form, although they are not actually
applied in exactly this form in the most well-known applications of
RBFs to 3D object recognition {\cite{PogEde90}}.

\Section{View-Based Model Acquisition}

Given that we have seen that a strongly two-dimensional observer can,
in fact, perform 3D object recognition as well as a Bayes-optimal 3D
model-based observer, we might ask the question of whether perhaps
view-based acquisition of new models requires more training in order
to achieve a comparable level of performance as direct,
coordinate-system based 3D model building and model-based recognition.

We have already answered that question implicitly in our derivation of
Bayes-optimal 3D recognition.  Bayes-optimal 3D recognition is carried
out in terms of (estimates of) $P(V|\mathcal{T}_\omega)$.  It makes no
difference how a vision system internally computes
$P(V|\mathcal{T}_\omega)$.  The computation may involved the
construction of explicit 3D object models, or it may be carried out in
some other way.  The computation may be carried out at the time when
the training views are first encountered, or it may be carried out
when the vision system is faced with the task of recognizing the
object represented by view $V$. All that matters is that the estimate
of $P(V|\mathcal{T}_\omega)$ ultimately is a good approximation to the
true value.

Since we have shown in the previous section that a strictly
two-dimensional observer can reconstruct the target and training views
perfectly from a set of real-valued similarity measurements, if that
observer chooses to evaluate $P(V|\mathcal{T}_\omega)$ by building a
3D model $M_\omega$ from training views internally (using techniques
like, e.g., \cite{higgins81}), the observer can simply do this in
terms of views reconstructed from the similarity measurements.

\Section{3D Model-Based Recognition}

In the previous sections, we have seen that strongly view-based
observers can perform Bayes-optimal 3D object recognition.  We also
showed that strongly view-based observers can perform model
acquisition as well as any 3D model-based recognition system.  In both
cases, the reason was that the set of similarity measurements
$\mathfrak{S}(V,\mathfrak{T})$ is essentially equivalent to complete
knowledge of all the training views and the target view.

Note that there is a distinction between Bayes-optimal 3D recognition
and 3D model-based recognition.  Bayes-optimal 3D recognition is
simply any procedure that achieves Bayes error rates on a 3D
recognition problem, regardless of what mechanisms it uses internally.
3D model-based recognition (at least in the sense used in this paper)
is based specifically on object-centered shape models.

Model-based 3D object recognition has been argued for in human vision
by Marr \cite{Marr82}, but work on 3D feature-based based object
recognition also usually assumes the existence of a 3D model (e.g.,
\cite{Grimson90z}).  Such models are usually assumed to be either
given (for example, from a CAD--computer aided design--model of the
object), or reconstructed from image data (e.g.,
\cite{higgins81,HerKan86,wells97}).

3D model-based recognition from collections of 2D training views
divides visual object recognition into two steps.  First, an
object-centered 3D shape model $\hat{M}_\omega$ is constructed based
on the training views $\mathcal{T}_\omega$.  Then, that 3D shape model
is used to find an match.

In its strictest form, this object centered shape model is a maximum
likelihood reconstruction or maximum a posteriori (MAP) reconstruction
$\hat{M}_\omega(\mathcal{T}_\omega)$ of the feature locations in 3D
from the set training views $\mathcal{T}_\omega$.
$\hat{M}_\omega(\mathcal{T}_\omega)$ is then used for performing
recognition.  If we assume that the 3D model match against the image
is carried out in a Bayes-optimal way, this means that we use
\begin{equation}\label{opt3dm}
  P ( V | \omega ) = \int P ( V | \hat{M}_{\omega}(\mathcal{T}_\omega), R, N ) P(R,N) dN dR dM_{\omega} \nonumber
\end{equation}
By comparing Equation~\ref{opt3dm} against Equation~\ref{opt3d}, we
see that this amounts to assuming that $P(M|\mathcal{T}_\omega) =
\delta(M,\hat{M}_\omega(\mathcal{T}))$.  This is correct
(and Bayes-optimal) when the object model is known exactly {\it a
priori}.  But when the object model has to be reconstructed from
training data, then, in general, $P(M|\mathcal{T}_\omega)$ is not
going to be a $\delta$ function.  The use of a maximum likelihood or
maximum a posteriori estimate for the model has to be justified as an
approximation; it is probably a good approximation when many training
views are available and/or the amount of noise is fairly small.

Therefore, model-based recognition using the ``best'' (in a maximum
likelihood sense) 3D model corresponding to the training views does
not necessarily lead to a Bayes-optimal 3D object recognition system.
To achieve Bayes-optimality, in general, it is necessary to model the
distribution $P(M|\mathcal{T}_\omega)$ correctly.

We can attempt to address this problem by adopting statistical 3D
shape models.  For example, we can associated each feature point in
the maximum likelihood or MAP reconstruction with error bounds or a Gaussian
distribution.  This, then, gives rise to a probability distribution
over possible 3D models compatible with the training views.  However,
this, too, only represents an approximation to the true distribution
$P(M|\mathcal{T}_\omega)$ because errors in the reconstruction of 3D
feature locations can (and usually are) correlated.

Overall, we see that using an object centered 3D shape model in 3D
model-based recognition, possibly with an associated error model, is
simply a particular choice of representation for
$P(M_\omega|\mathcal{T}_\omega)$.  But we have seen such uses of 3D
models in recognition correspond to specific assumptions about
$P(M_\omega|\mathcal{T}_\omega)$, assumptions that may not be
satisfied in specific recognition problems.  Or, to put it more
succinctly, combining optimal 3D model reconstruction from training
views with optimal 3D model matching against 2D images does not
necessarily result in Bayes-optimal 3D recognition.  

\Section{Discussion}

A key result of this paper is that a strongly two-dimensional
observer, that is, an observer that performs object recognition only
in terms of Euclidean similarity measures between different views, can
achieve the same Bayes-optimal performance as an observer having full
knowledge of all the geometric information contained within views.
The reason was that the strongly two-dimensional observer has all the
information necessary to reconstruct the essential geometric
information contained in the views: strongly view-based recognition is
really nothing more than a change of coordinate system in which visual
input is represented.  And while we used the concrete example of
objects consisting of point-like features, as used in prior work in
the literature, the same approach works for many other forms of view
representations, for example, in terms of locations without known
correspondences or gray-value pixel values.

As a consequence, it is impossible to distinguish definitively 3D model-based
recognition from strongly view-based recognition by comparing the error rates
of different observers: both 3D model-based observers and view-based observers
can achieve the same Bayes-optimal 3D recognition and model acquisition
performance; either of them may fall short if the observer is using a
suboptimal implementation. 

These results seem to be in contradiction to those claimed in
{\cite{LiuKniKer95}}{\cite{LiuKer98}}.  In those papers, the authors define  ``ideal''
2D observers and demonstrate that human performance and 3D model-based
recognition exceeds that of those ideal observers.  However, while those
papers compare human performance to some 2D observers (and, in fact, observers
that are Bayes-optimal for certain 2D matching problems {\cite{wells97}}), the
2D observers in those papers simply are not the best possible that can be
constructed with 2D similarity methods and arbitrary combination of evidence
procedures.

Whether any meaningful and testable hypotheses distinguishing view-based and
3D model-based recognition systems and strategies can be formulated at all
remains to be seen.  It might be useful to shift the debate from
considerations of what operations are involved in the recognition of
individual objects to the prior knowledge about the world that a 3D
model-based system is created with.  A Bayes-optimal 3D model-based system
should be able to perform perfect view generalization without any training,
while a more general-purpose visual recognition system would require time to
learn the view generalization function.  On the other hand, a Bayes-optimal 3D
model-based system might not be able to adapt well to objects whose appearance
transforms in ways other than that expected of 3D models under changes in
viewing position \cite{obscured-breuel92phd}.  However, experimentation in these
areas is difficult because ``training'' refers to the entire visual experience
of a human observer throughout his life, not to the acquisition of individual
object models.

In fact, the considerations in the last section have shown that 3D
model-based recognition systems that either just perform a maximum
likelihood or MAP reconstruction of a 3D model from training views, or even
systems that associate error bounds with such reconstructions, are not
Bayes optimal for 3D recognition in general.  Bayes optimal
recognition in general requires correct modeling of the distribution
$P(M_\omega|\mathcal{T}_\omega)$, and approximating that distribution
well under the constraint that it be represented in terms of
perturbations of a concrete 3D shape model may be very difficult and,
in any case, is not usually attempted by 3D model-based recognition
systems anyway.  View-based models, instead, attempt to model
$P(V|\omega)$ or $P(V|\mathcal{T}_\omega)$ directly without imposing
the constraint that the representation of that density be tied somehow
to a 3D shape model.  Whether this is actually easier or more
successful in practice remains to be seen, but it is certainly a valid
alternative to 3D shape models, and it allows us to explore a much
larger space of possible probabilistic models.

The reconstruction methods used in this paper are a mathematical device to
establish statistical sufficiency.  While reconstruction from distances could
probably be accomplished by simple constraint propagation in hardware that
might plausibly described as ``neural'', this is entirely unnecessary.  Any
classification method that achieves Bayes-optimal asymptotic performance given
enough training data would be expected eventually learn the view generalization
function, whether it is expressed in terms of Euclidean distances to prototype
views or in terms of coordinates.  The coordinate transformation implied by
view-based representations, using distances to prototype views, does not seem
particularly complex and might even simplify the learning problem for class
conditional densities or view generalization functions.  Therefore, we should
not judge the plausibility of Bayes-optimal view-based recognition in an
actual vision system by the mathematical techniques used in this paper for
establishing statistical sufficiency.  The question of whether we can
construct Bayes-optimal view generalization functions that are based on
strongly two-dimensional techniques is a question of complexity, as well as
the distribution of actual shapes and views in the real world, and will be
addressed in a separate paper.

\bibliographystyle{plain}
\bibliography{iccv.bib}

\begin{thebibliography}{10}

\bibitem{belongie01shape}
S.~Belongie, J.~Malik, and J.~Puzicha.
\newblock Shape matching and object recognition using shape contexts, 2001.

\bibitem{Grimson90z}
Eric Grimson.
\newblock {\em {Object Recognition by Computer}}.
\newblock MIT Press, Cambridge, MA, 1990.

\bibitem{HerKan86}
M.~Herman and T.~Kanade.
\newblock Incremental reconstruction of {3D} scenes from multiple, complex
  images.
\newblock {\em Artificial Intelligence}, 30(3):289--341, 1986.

\bibitem{wells97}
William~Wells {III}.
\newblock Statistical approaches to feature-based object recognition.
\newblock {\em International Journal of Computer Vision}, 21(1/2):63--98, 1997.

\bibitem{LiuKer98}
Z.~Liu and D.~Kersten.
\newblock 2d observers for human {3D} object recognition?
\newblock {\em Vision Research}, 38:2507--2519, 1998.

\bibitem{LiuKniKer95}
Z.~Liu, D.~C. Knill, and D.~Kersten.
\newblock Object classification for human and ideal observers.
\newblock {\em Vision Research}, 35(4):549--568, 1995.

\bibitem{higgins81}
Christopher Longuet-Higgins.
\newblock A computer algorithm for reconstructing a scene from two projections.
\newblock {\em Nature}, 293:133--135, 1981.

\bibitem{Marr82}
David Marr.
\newblock {\em {Vision: A Computational Investigation into the Human
  Representation and Processing of Visual Information}}.
\newblock W.H. Freeman and Company, San Francisco, 1982.

\bibitem{PogEde90}
T.~Poggio and S.~Edelman.
\newblock A network that learns to recognize three-dimensional objects.
\newblock {\em Nature}, 343:263--266, 1990.

\bibitem{UllBas91}
S.~Ullman and R.~Basri.
\newblock Recognition by linear combination of models.
\newblock {\em IEEE Trans. PAMI}, 13:992--1006, 1991.

\bibitem{Helmholtz25}
H.~von Helmholtz.
\newblock {\em Treatise on physiological optics: the direction of vision}.
\newblock Optical Society of America, Rochester, New York, 1925.

\end{thebibliography}

\section*{Appendix A}

Here is a brief sketch of the proof of Lemma \ref{reconstruction}:

\begin{lemma}
  The intersection of two hyperspheres $A = \{ x \in \mathbb{R}^n | ( x - a
  )^2 = r^2_a \}$ and $B = \{ x \in \mathbb{R}^n | ( x - b ) = r^2_b \}$ of
  dimension n-1 is either empty, a single point, an $n - 2$ dimensional
  hypersphere contained in an $n - 1$ dimensional linear subspace
  perpendicular to $( b - a )$, or $A = B$. \label{hysect}
\end{lemma}

Assume we are given $A$ and $B$.  If $a = b$ and $r_a = r_b$, then $A = B$. 
If $a = b$ and $r_a \neq r_b$, then the intersection is empty. Therefore, let
us assume that $a \neq b$ and that there is a common point $p \in A, B$. 
Without loss of generality, place $a$ at the origin, $a = 0$.  Write $p =
\lambda ( b - a ) + q = \lambda b + q$, where $b \cdot q = 0$.  Plugging this
into the equations for $A$ and $B$, we obtain $\lambda^2 + q^2 = r_a^2$ and $(
1 - \lambda )^2 + q^2 = r_b^2$.  Solving for $\lambda$ yields $\lambda =
\frac{1}{2 | | b | |^2} ( r_a^2 - r_b^2 + | | b | | ) + 1$ and $| | q | | =
\sqrt{r_a^2 - \lambda^2}$, which establishes the claim. $\Box$

\begin{lemma}
  For a collection of n linearly independent vectors $p_1, \ldots, p_n$ in
  $\mathbb{R}^n$, we can reconstruct the coordinates of the vectors  from the
  collection of Euclidean distances $d_{i j} = \| p_i - p_j \|$ up to a global
  translation, a global rotation, and mirror image
  reversal.
\end{lemma}

If $n = 1$, we have a single point, which we place at the origin, giving us a
solution up to translation. Now, take distances $d_{i j}$ for $i, j \leq n -
1$ and apply the Lemma, giving a collection of points $p_1, \ldots, p_{n - 1}
\in \mathbb{R}^{n - 1}$. Map that solution into $\mathbb{R}^n$ by adding $0$
as the last coordinate to each vector; this corresponds to an
arbitrary choice of rotation.  Now consider the hyperspheres around
each point $p_i$ with radius $d_{n i}$.  By Lemma \ref{hysect}, their
intersection will be a linear subspace of dimension 1, containing a
hypersphere of dimension 0, i.e., two points.  It is left to the
reader to prove that these are mirror symmetric around the plane $\{ v
\in \mathbb{R}^n | v^n = 0 \}$. $\Box$

\textit{\textup{\textbf{Lemma~\ref{reconstruction}.}} For a collection of N distinct vectors
$p_1, \ldots, p_N$ that span $\mathbb{R}^n$, if $N > n$, we can reconstruct
the coordinates of the vectors from the collection of Euclidean distances
$d_{i j} = \| p_i - p_j \|$ up to a global translation, a global rotation, and
mirror reversal.}

Find a linearly independent subset of $n$ vectors and apply Lemma
\ref{reconstruction}, giving $p_1, \ldots, p_n$.  Now, consider the
reconstruction of $p_q = p_{n + 1}, \ldots, p_N$.  Place spheres of
radius $d_{q i}$ around each $p_i$, $i = 1, \ldots, n$ and compute
$p_q$ as the intersection of the linear subspaces from Lemma
\ref{hysect}.  The reader can prove for himself that seen that this
intersection has to be unique. $\Box$

\section*{Appendix B}

In this Appendix, we construct a similarity function $\mu$ that permits exact
reconstruction of $V$ and $T_{\omega, i}$ given only the values of $\mu ( V,
T_{\omega, i} )$ for a single $\omega$.  This is an alternative construction
to that given in the text, which potentially required knowledge of the
similarity of a target view to the training views for multiple objects
$\omega$ and reconstructed views only up to a global translation, rotation,
and mirror image.

\begin{theorem}
  There exists a real-valued function $\mu : \mathbb{R}^{2 k} \times
  \mathbb{R}^{2 k} \rightarrow \mathbb{R}$ and a function $f : \mathbb{R}^n
  \rightarrow \mathbb{R}$ such that $P ( V | T_1, \ldots, T_r ) = f ( \mu ( V,
  T_1 ), \ldots, \mu ( V, T_r ))$.
\end{theorem}

Here, $\mu$ is the ``view similarity function'' and $f$ is the ``combination
of evidence function''.

For the proof of this theorem, we require a family of functions (one for each
value of $k$) $\iota : \mathbb{R}^k \rightarrow \mathbb{R}$ and its inverse
$\iota^{- 1} : \mathbb{R} \rightarrow \mathbb{R}^k$ such that $\iota^{- 1} (
\iota ( x )) = x$ for any $x$ in $\mathbb{R}^k$. We can construct a function
$\iota$ easily by interleaving the digits of the individual arguments. That
is, let $x_i = \sum_{j = - \infty}^{\infty} d_{i j} 10^j$. Then, $\iota ( x )
= \sum_{j = - \infty}^{\infty} d_{j \tmop{div} k, j \tmop{mod} k} 10^j$ If $x'
= \iota ( x ) = \sum_{j = - \infty}^{\infty} d'_j 10^j$, then $x_i = \sum_{j =
- \infty}^{\infty} d'_{j k + i} 10^j$.

Now, let $v = ( S, T_i )$ be the concatenation of the vectors $S$ and $T_i$
and let $v_S$ and $v_T$ denote the portions of the vector $v$ corresponding to
$S$ and $T$ respectively in such a concatenation. Choose $\mu ( S, T_i ) =
\iota (( S, T_i ))$ and choose $f_{} ( \mu_1, \ldots, \mu_r ) = P ( S | T_1,
\ldots, T_r ) = P ( \iota^{- 1} ( \mu_1 )_S | \iota^{- 1} ( \mu_1 )_T, \ldots,
\iota^{- 1} ( \mu_r )_T, \ldots )$. By construction, $f_{} ( \mu ( S, T_1 ),
\ldots, \mu ( S, T_r )) = P ( S | T_1, \ldots, T_r )$. We have therefore shown
that any Bayes-optimal decision function based on 3D models can be expressed
as a decision function involving only real-valued similarity functions $\mu (
S, T_i )$ and the combinations of the similarity scores. Note that in this
construction $f_{}$ is not even object-dependent. $\Box$

While the function $\iota$ used in this construction happens to be not
continuous, a construction using a Hilbert curve (space filling curve) for
$\iota$ would allow us to derive essentially the same result.

\end{document}